\documentclass[sigconf, nonacm]{acmart}

\usepackage{tcolorbox} 
\usepackage{mdframed} 
\usepackage[shortlabels]{enumitem}
\usepackage{booktabs}
\usepackage{arydshln}
\usepackage{multirow}

\AtBeginDocument{%
  }

\setcopyright{acmlicensed}
\copyrightyear{2026}
\acmYear{2026}
\acmDOI{XXXXXXX.XXXXXXX}
\acmConference[ICAIL 2026]{...}{June 09--12, 2026}{Singapore}
\acmISBN{XXX-X-XXXXX-XXXX-X/XXXX/XX}

\setcopyright{none}
\renewcommand\footnotetextcopyrightpermission[1]{}

\begin{document}

\newcommand{\researchquestionbox}[1]{
    {
    \vspace{-1.2mm}
    \begin{tcolorbox}[colback=white!0,arc=0.65mm,boxrule=1pt,right=2mm,left=2mm,bottom=2mm,top=2mm]
        \fontsize{10pt}{10pt}\selectfont
        \vspace{-1mm} #1
        
        \vspace{-2mm}
    \end{tcolorbox}
    }
    \vspace{-2mm}
}

\newcommand{\researchquestion}[2]{
    {
    \vspace{-1.2mm}
    \begin{tcolorbox}[colback=white!0,boxrule=0.9pt,right=2mm,left=2mm,bottom=2mm,top=2mm]
        \fontsize{10pt}{10pt}\selectfont
        \vspace{-1mm} \textbf{#1}: #2
        
        \vspace{-0.8mm}
    \end{tcolorbox}
    }
    \vspace{-2mm}
}

\title{Legal Domain Adaptation of Modern BERT Models}

\author{Dominik Stammbach}
\email{dominsta@princeton.edu}
\orcid{0000-0003-1631-3020}

\affiliation{%
  \institution{Princeton University}
  \city{Princeton}
  \state{New Jersey}
  \country{USA}
}

\author{Peter Henderson}
\email{peter.henderson@princeton.edu}

\affiliation{%
  \institution{Princeton University}
  \city{Princeton}
  \state{New Jersey}
  \country{USA}
}

\renewcommand{\shortauthors}{Stammbach and Henderson}

\begin{abstract}
We investigate domain adaptation of modern BERT models in the legal domain. We further pre-train ModernBERT on all US court opinions using the masked language modeling objective. Although ModernBERT has been trained on roughly 500x more data than original BERT, we still find that this model benefits from further pre-training and domain adaptation in the legal domain: we report significant improvements compared to vanilla ModernBERT on all datasets connected to US court opinions. We find gains similar to those reported in early work on domain adaptation of BERT-like models. However, from scratch pre-training does not match the performance of further pre-training an existing ModernBERT checkpoint in our experiments. The resulting models are capable of processing sequences up to 8,192 tokens, and can be used to compute meaningful embeddings of legal passages, or could quickly rerank hundreds of legal passages for a given search query. We release all model checkpoints publicly.
\end{abstract}


\maketitle

\section{Introduction}

There has been a resurgence in research on BERT-style encoders \citep{portes2023MosaicBERT, modernbert}. While interest over the last few years predominantly shifted towards generative AI, BERT-style encoders throughout remain some of the most widely used NLP models in practice \citep{modernbert}. These newer encoders, especially ModernBERT, implement some of the improvements which are fairly standard in current GPT-style models, e.g., flash attention \cite{dao2023flashattention2fasterattentionbetter}, RoPE embeddings \cite{SU2024127063} and training on modern trillion-token data scales, similar to recent GPT-style models.

Given that ModernBERT has been trained on much more data, we ask whether these encoders still profit from domain adaptation. We start with ModernBERT \cite{modernbert}, a BERT-style encoder which has been trained on two trillion tokens. This is a roughly 500x increase compared to the original BERT model \citep{bert}. First, we find that further masked language modeling pre-training on case law data significantly improves performance on NLP tasks related to US court opinions and summarize our main results in Table \ref{tab:summary_results}. We present statistically significant improvements (Table \ref{tab:main_results}) over a vanilla ModernBERT baseline on all NLP tasks concerned with US court opinions \cite{lexglue, mahari-etal-2024-lepard, Zheng_2025}. Therefore, we speculate that ModernBERT has not been widely pre-trained on legal or caselaw data\footnote{The paper does not disclose details about the pre-training corpus.} and release the resulting model checkpoints.\footnote{{Model checkpoints are available on Hugging Face: 
\href{https://huggingface.co/ai-law-society-lab/CaseLawModernBERT-base}{https://huggingface.co/ai-law-society-lab/CaseLawModernBERT-base} and 
\href{https://huggingface.co/ai-law-society-lab/CaseLawModernBERT-large}{https://huggingface.co/ai-law-society-lab/CaseLawModernBERT-large}.}}

\begin{table}[t!]
  \centering

  \caption{Summary of main results}
  \label{tab:summary_results}

  \footnotesize
  \begin{tabular}{lccccc}
    \toprule
    Model & SCOTUS & CaseHold & LePaRD & BarExam QA & avg \\
    \midrule
    ModernBERT & 65.3 & 74.8 & 70.6 & 2.6 & 53.1 \\
    Legal ModernBERT &
    \textbf{67.5} & \textbf{76.1} & \textbf{71.5} & \textbf{5.3} & \textbf{55.1} \\
    \bottomrule
  \end{tabular}

  \vspace{5pt}
  \footnotesize
  \emph{We report the average over five runs with different seeds for all experiments. Results are statistically significant, see Table \ref{tab:main_results} for full results and 95\% confidence intervals.} 
\end{table}

Second, and more relevant to the broader NLP community, we find that pre-training a model with the ModernBERT architecture from scratch with a customized in-domain vocabulary does not match the performance of further pre-training an existing ModernBERT checkpoint. This is in contrast to earlier work on in-domain pre-training where this strategy consistently achieved best results \citep{beltagy-etal-2019-scibert, chalkidis-etal-2020-legal, zheng2021doespretraininghelpassessing, webersinke2022climatebertpretrainedlanguagemodel}. We do not rule out the possibility that larger-scale pre-training from scratch eventually matches or surpasses the performance of our best model. However, we do not observe this when we pre-train on all US court opinions.

\begin{table*}[h!]
    \centering
    
    \caption{Test set results on two LexGLUE tasks. We report the average across 5 runs with 95\% confidence intervals. As a competitive baseline, we choose the best reported model in \citep{chalkidis-etal-2020-legal}.
    ** denotes statistically significant results.}
    \label{tab:main_results}
    
    \small
    \begin{tabular}{l l c c }
        Dataset & Model & Micro-F1 & Macro-F1 \\ \hline
        \multirow{4}{*}{LexGLUE/SCOTUS base} & Best Reported in \cite{lexglue} & 65.9 & 76.6 \\ 
        & ModernBERT-base & 65.3 $\pm$ 0.8 & 75.7 $\pm$ 0.6 \\
        & LegalModernBERT-base from scratch & 63.4 $\pm$ 0.9 & 75.2 $\pm$ 0.8 \\
        & LegalModernBERT-base & \textbf{67.5 $\pm$ 0.9**} & \textbf{77.5 $\pm$ 0.7**} \\ \hdashline
        \multirow{4}{*}{LexGLUE/SCOTUS large} &  RoBERTa-large \cite{lexglue} & 66.3 & 75.5 \\
        & ModernBERT-large & 68.6 $\pm$ 1.2 & 77.9 $\pm$ 0.9 \\
        & LegalModernBERT-large from scratch & 63.4 $\pm$ 0.9 & 75.2 $\pm$ 0.8 \\
        & LegalModernBERT-large & \textbf{69.3 $\pm$ 1.1} & \textbf{78.5} $\pm$ \textbf{0.7} \\ \hline
        \multirow{4}{*}{LexGLUE/CaseHold base} & Best Reported in \cite{lexglue} & 75.4 & 75.4 \\ 
        & ModernBERT & 74.8 $\pm$ 0.3 & 74.8 $\pm$ 0.3 \\
        & LegalModernBERT from scratch & 74.9 $\pm$ 0.4 & 74.9 $\pm$ 0.4 \\
        & LegalModernBERT & \textbf{76.1 $\pm$ 0.2**} & \textbf{76.1 $\pm$ 0.2**} \\ \hdashline
        \multirow{4}{*}{LexGLUE/CaseHold large} & RoBERTa-large \cite{lexglue} & 74.4 & 74.4 \\ 
        & ModernBERT-large & 76.9 $\pm$ 0.7 & 76.9 $\pm$ 0.7 \\
        & LegalModernBERT-large from scratch & 75.0 $\pm$ 0.7 & 75.0 $\pm$ 0.7 \\
        & LegalModernBERT-large & \textbf{78.6 $\pm$ 0.1**} & \textbf{78.6 $\pm$ 0.1**} \\ \hline
        
    \end{tabular}

\end{table*}

Previously, \citet{lexglue} discussed that future advances in legal NLP require the ability to process longer texts, recognize document structure, more large-scale pre-training, and larger models. Our work speaks to all these points: LegalModernBERT is able to process documents of up to 8,192 tokens. This is longer than the length of the average US court opinion \cite{Feldman2018}, and enables to compute full court opinions, which was impossible with the 512 token sequence length of prior BERT models. Also, training on long documents at scale implicitly picks up document structure \citep{grattafiori2024llama3herdmodels}. Additionally, LegalModernBERT has undergone large-scale pre-training, and in total was trained on 2 trillion tokens of general language and an additional 13 billion tokens of long document court opinions. To address larger models, we additionally further pre-train large model checkpoints on court opinions. Perhaps unsurprisingly, we observe consistent improvements for the LegalModernBERT-large models throughout all experiments on all benchmarks.

Similar to \citep{modernbert}, we believe there is practical value of BERT-style encoders, especially in the legal domain. Tasks in that domain typically consist of embedding and retrieving documents, reranking, anonymizing or finding relevant evidence in large amounts of data \citep[e.g.,][]{zhong-etal-2020-nlp, mahari-etal-2023-law, cheong2026aiaugmentaccessjustice}. In practice, legal data is especially sensitive.
\citet{cheong2026aiaugmentaccessjustice} discuss legal AI applications for public defenders, and note that data submitted to commercial proprietary models could fall outside of privileged information and be subject to mandatory disclosure~\cite{privilege, cheong2024not}. Thus, one pathway for public defender AI systems are closed universe, deployed in-house, and relying on small, specialized open source models \citep{cheong2026aiaugmentaccessjustice}. 

Our work connects to the goal of increasing access to justice with the help of NLP, as articulated in \cite{mahari-etal-2023-law, cheong2026aiaugmentaccessjustice, stammbach2026legalretrievalpublicdefenders}. All of the tasks mentioned above, most notably embedding, retrieving and reranking passages, and finding relevant evidence in vast amounts of documents, could directly benefit from more capable, domain-adapted modern BERT-style models. We believe the released models in this work are a first step towards that goal. To summarize, our contributions are:

\researchquestionbox{
\begin{enumerate}[(1),left=0mm,topsep=0.1mm,noitemsep]
\item We investigate domain adaptation of modern BERT encoders.
\item We release a number of legal ModernBERT models capable of processing long sequences, consistently outperforming the vanilla checkpoints in tasks related to US court opinions. 
\end{enumerate}
}

\section{Related Work}

Since the introduction of BERT \cite{bert}, researchers have been investigating the domain adaptation of such models. In the legal domain, related work includes LEGAL-BERT \citep{chalkidis-etal-2020-legal}, CaseLawBERT \citep{zheng2021doespretraininghelpassessing} and PileOfLaw-BERT \cite{hendersonkrass2022pileoflaw}. All these have used methods introduced in the original BERT paper \citep{bert} or RoBERTa \cite{roberta}. Moreover, BERT checkpoints have been adapted to many different domains, for example scientific text \citep{beltagy-etal-2019-scibert} or environmental text \citep{webersinke2022climatebertpretrainedlanguagemodel}. 

BERT-style encoders have been very resilient over the years, and especially the RoBERTa-large \citep{roberta} remains a tough-to-beat baseline for encoder-only architectures. However, some of the advances discovered in GPT-style models have recently been incorporated in BERT-like models. \citet{portes2023MosaicBERT} introduce MosaicBERT with flash attention and ROPE embeddings, capable of processing longer sequences efficiently. And finally, \citet{modernbert} introduce ModernBERT, similar to MosaicBERT, but pre-trained on a modern data scale of 2 trillion tokens.

To the best of our knowledge, domain adaptations of these recent BERT architectures for the legal domain \citep{portes2023MosaicBERT, modernbert} have not been explored in much detail yet. Thus, we investigate the performance of ModernBERT in the legal domain, and compare to checkpoints which are further pre-trained on in-domain data. Our findings overall remain similar to early findings in domain adaptation of BERT models: \textbf{Continued pre-training helps}, and \textbf{improvements} in the range of 1-3 percentage points increase on evaluated tasks \textbf{are similar to early domain adaptation results} \citep{chalkidis-etal-2020-legal, zheng2021doespretraininghelpassessing, hendersonkrass2022pileoflaw}.

\begin{table*}[]
    \centering
    \caption{Results on retrieval tasks using a fine-tuned bi-encoder following \citep{reimers-2019-sentence-bert}. Results are the average of 5 runs with 95\% confidence intervals. ** denotes statistically significant results.}
    \label{tab:my_label}
    \small
    \begin{tabular}{l c c | c c}
    Model & \multicolumn{2}{c}{LePaRD} &  \multicolumn{2}{c}{BarExam QA}   \\
    & rc@1 & rc@10 & rc@1 & rc@10 \\ \hline
    ModernBERT-base & 24.58 $\pm$ 0.1 & 70.64 $\pm$ 0.1 & 0.48 $\pm$ 0.6 & 2.58 $\pm$ 0.3 \\
    LegalModernBERT-base from scratch & \textbf{25.33 $\pm$ 0.0**} & \textbf{72.22 $\pm$ 0.0**} & \textbf{2.9 $\pm$ 0.3**} & \textbf{6.45 $\pm$ 0.4} \\
    LegalModernBERT-base & 24.95 $\pm$ 0.1 & 71.46 $\pm$ 0.0 & 0.97 $\pm$ 0.3 & 5.32 $\pm$ 0.6
 \\ \hdashline
    ModernBERT-large & 26.7 $\pm$ 0.1 & 73.93 $\pm$ 0.1 & 1.61 $\pm$ 0.0 & 5.65 $\pm$ 0.4 \\
    LegalModernBERT-large from scratch & 26.0 $\pm$ 0.0 & 73.39 $\pm$ 0.0 & \textbf{4.84 $\pm$ 0.0**} & 10.48 $\pm$ 0.0
 \\
    LegalModernBERT-large & \textbf{27.06 $\pm$ 0.1**} & \textbf{74.75 $\pm$ 0.1**} & 3.23 $\pm$ 1.0 & \textbf{11.77 $\pm$ 0.8**} \\ \hline
    \end{tabular}

\end{table*}

\section{Experiments}

We conduct four pre-training experiments: First, we further pre-train existing ModernBERT checkpoints (base and large) on all 8.3 million US court opinions (13B words) found in the Collaborative Open Legal Data \citep{harvard-lil_cold_cases}. Second, we pre-train models (base and large) from scratch on the same data using an in-domain legal vocabulary taken from \citep{zheng2021doespretraininghelpassessing}.

We also experimented with a cleaned version of the COLD cases \cite{ccap}. We pre-trained two models, one on 5\% of the original COLD cases, and one on 5\% of the cleaned cases. We found that the performance for the model trained on the original data was slightly higher on the LexGLUE/SCOTUS dataset. Moreover, we did not find enough details about the exact cleaning steps in \cite{ccap}. Thus, we decided to proceed with the original COLD case data.

We run all experiments with the same hyper-parameters found in the ModernBERT paper: we use a masked ratio of 30\%, a starting learning rate of 3e-4, a linear LR scheduler and weight decay (8e-5). We opt for a constant batch size of 512 where each example is up to 8,192 tokens long. We discard opinions shorter than 500 tokens, split opinions longer than 8,192 tokens into multiple examples, and pad opinions longer than 500 tokens, but shorter than 8,192 token. Each example contains tokens from a single court opinion only. We display pre-training loss curves of four pre-training runs (base and large, pre-training from existing checkpoint / from scratch) in Appendix \ref{appendix:model_training}.

\section{Results}\label{sec:results}

We evaluate all models on four tasks related to US court opinions: LexGLUE/SCOTUS \cite{spaeth2020supreme}, LexGLUE/CaseHold \citep{zheng2021doespretraininghelpassessing}, LePaRD \cite{mahari-etal-2024-lepard} and BarExam QA \citep{Zheng_2025}. The goal of LexGLUE/SCOTUS is to classify a court opinion into 14 issue areas \citep{spaeth2020supreme}.
LexGLUE/CaseHold is a multiple choice QA task where the model has to identify the correct holding \citep{zheng2021doespretraininghelpassessing}. LePaRD is a legal passage retrieval task with the goal to identify the correct cited passage given some preceding context \citep{mahari-etal-2024-lepard}. In BarExam QA, Bar Exam hypotheticals are manually linked to a paragraph connected to the hypothetical \citep{Zheng_2025}. The task then is to retrieve the relevant paragraph. All experiments are run with the hyper-parameters found in the corresponding replication packages of these benchmarks.

Overall, we find consistent results across all experiments. In both LexGLUE tasks, ModernBERT-base outperforms most non-specialized BERT checkpoints reported in \citep{lexglue}, but does not match the performance of early in-domain legal BERT variants. Additionally, legal checkpoints pre-trained from scratch do not match the performance of a vanilla ModernBERT checkpoint. However, further pre-training a ModernBERT checkpoint on in-domain data yields statistically significant improvements on both these tasks in three out of four experiments. In the LexGLUE/SCOTUS-large setting, we observe high variance and thus large confidence intervals. These findings replicate on the development set and we show the corresponding results in Appendix Table \ref{tab:dev_main_results}. 

Surprisingly, pre-training a model with the ModernBERT architecture from scratch with an in-domain vocabulary does not reach the performance of a further pre-trained LegalModernBERT checkpoints. This is in contrast to earlier related work on domain adaptation, where this approach yielded (somewhat consistently) best results \citep{chalkidis-etal-2020-legal, zheng2021doespretraininghelpassessing}. We speculate that pre-training on 2 Trillion tokens offsets the gains made possible by from-scratch pre-training (as can be seen in the loss curves in Appendix \ref{appendix:model_training}). Second, the vocabulary of ModernBERT is more adaptive towards new domains than the vocabularies used in e.g., BERT or RoBERTa. We tokenized all training set examples in SCOTUS and find that the average sequence length of the in-domain tokenizer results in on average 7,998 tokens per example. Examples tokenized with ModernBERT are 1.9\% longer; examples tokenized with RoBERTa are 3.7\% longer. However, we note that these findings might change if we were to train models from scratch on more large-scale in-domain data, e.g., all data in The Pile of Law \citep{hendersonkrass2022pileoflaw}.

\begin{table*}[t!]
    \centering
    \caption{Ablation experiment for different sequence lengths. Performance on LexGLUE/SCOTUS, once trained with a maximum sequence length of 8,192, compared to training with a sequence length of 512. In brackets, performance gains if trained with larger sequence length.}
    \label{tab:ablation_seq_length}
    \small
    \begin{tabular}{p{5cm} l c c }
        Dataset & Model & Micro-F1 & Macro-F1 \\ \hline
        \multirow{6}{*}{\parbox[c]{5cm}{LexGLUE/SCOTUS (512 sequence length)}} &  ModernBERT-base & 56.8 & 70.3  \\
        & LegalModernBERT-base from scratch & 60.4 & 73.2 \\
        & LegalModernBERT-base & 61.5  & 73.6 \\ 
        &  ModernBERT-large & 63.0 & 72.7 \\
        & LegalModernBERT-large from scratch & 62.1 & 71.7 \\
        & LegalModernBERT-large & 67.0 & 76.8 \\ \hdashline
        \multirow{6}{*}{\parbox[c]{5cm}{LexGLUE/SCOTUS (8,192 sequence length)}}        
        &  ModernBERT-base & 65.3 (+8.5) & 75.7 (+ 5.4) \\
        & LegalModernBERT-base from scratch & 63.4 (+ 3.0) & 75.2 (+ 2.0) \\
        & LegalModernBERT-base & 67.5 (+ 6.0) & 77.5 (+3.9) \\ 
        & ModernBERT-large & 68.6 (+5.6) & 77.9 (+5.2) \\
        & LegalModernBERT-large from scratch & 63.4 (+ 1.3) & 75.2 (+ 3.5) \\
        & LegalModernBERT-large & \textbf{69.3} (+ 2.3) & \textbf{78.5} (+ 1.7) \\ \hline

    \end{tabular}

\end{table*}

In Table \ref{tab:ablation_seq_length}, we show that the longer sequence length of ModernBERT and our domain-adapted variants make a difference: In the four examined datasets, only SCOTUS sequences are consistently longer than 512 tokens (the maximum sequence length of the original BERT model). We show that fine-tuning on a sequence length of 8,192 tokens always increases performance in all experiments by a substantial margin. We interpret this finding as evidence that legal NLP using bidirectional encoders so far was limited by models with insufficiently large context lengths.

For the retrieval-oriented tasks, we again find that domain-adapted models obtain better results than vanilla ModernBERT. Sometimes, we obtain the best results with from scratch pre-training: we believe that retrieval tasks reward lexical specialization more than classification tasks, especially for smaller models. Thus, the legal base model from scratch with a specialized vocabulary seems to perform best among base models, while the LegalModernBERT-large checkpoint still obtains the best results overall among large models. 

These findings all replicate in zero-shot experiments where we do not fine-tune on the LePaRD training set. In-domain variants already contain a more similar representation of queries and targets in the LePaRD dataset (see Table \ref{tab:zero-shot-lepard}. Furthermore, \citet{mahari-etal-2024-lepard} conduct experiments where retrieval is perceived as classification \citep{tay2022transformermemorydifferentiablesearch}. We analogously implement this experimental setup and report results in Appendix Table \ref{tab:lepard_classification} -- again, LegalModernBERT-large consistently yields the best results.

Lastly, we conduct probing experiments (assessing model knowledge via cloze queries) following \citep{chalkidis-etal-2023-lexfiles}. We present results in Appendix Table \ref{tab:probing_results}. Our LegalModernBERT-large checkpoint achieves the highest P@1 and MRR for US terms and US crimes -- and for the average score. At the same time, we observe that P@1 and MRR for e.g., European Court of Human Rights terms decrease. We believe this is due to our continued pre-training only on US court opinions, which likely are a different distribution than ECHR terms.

\section{Conclusion}\label{sec:conclusion}

In this work, we investigate legal domain adaptation of more recent BERT encoders, specifically ModernBERT \cite{modernbert}. To the best of our knowledge, this is the first work to do so. Similar to early work in domain adaptation, we find that legal domain adaptation helps ModernBERT models as well. Effect sizes are consistent with prior work on domain adaptation. 

As part of this work, we release all domain-adapted LegalModernBERT checkpoints, which can be useful for NLP tasks related to US court opinions, or developing closed universe legal tech with specialized in-house models. Our recommendation is to start with the released LegalModernBERT-large checkpoint, which consistently obtained strongest results on various classification, multiple-choice QA and retrieval tasks in all our experiments.
\vspace{-0.2em}

\newpage
\section*{Limitations}

\paragraph{Work limited to BERT-models.} This work is about modern BERT-style encoder models \citep{modernbert} and their domain adaptation to US court opinions. Thus, the paper does not engage with other advances in legal NLP, for example the potential of GPT-4 to generate accessible summaries of court opinions \cite{translating_legalese}, domain adaptation of Llama models \citep{dominguezolmedo2024lawmapowerspecializationlegal}, or legal instruction fine-tuning of T5 models \citep{niklaus2025lawinstructresourcestudyinglanguage}. 

\paragraph{Work limited to US court opinions, although other legal data sources exist.}\vspace{-0.2em} We only investigate pre-training on US court opinions \cite{harvard-lil_cold_cases}, and do not engage with other sources of legal textual data \citep[e.g.,][]{hendersonkrass2022pileoflaw, niklaus2024multilegalpile689gbmultilinguallegal}. We suspect similar results to LegalBERT \cite{chalkidis-etal-2020-legal} or other similar work if we were to fine-tune on more diverse sources of legal text, and leave it to future work to explore these in more detail.

\paragraph{Model performance on other legal NLP tasks.}\vspace{-0.2em} We only evaluate our model on legal NLP tasks connected to US court opinions, specifically LexGLUE/SCOTUS, LexGLUE/CaseHold  and LePaRD. While it would be interesting to explore performance on other legal NLP tasks, such as predicting judicial outcomes \citep{lexglue} or others, we think performance likely suffers on these tasks as they're not closely connected to our pre-training data. Furthermore, such a study would warrant pre-training on more diverse sources of legal data.

\paragraph{Noise in the COLD cases.}\vspace{-0.2em} Opinions are usually published in a PDF format. These PDFs are then converted into text, and the resulting text can contain errors typical in such conversion efforts at scale. We did not attempt to clean the data and only applied minimal preprocessing. As already discussed in the paper, we experimented with a cleaned version of the court opinions \citep{ccap}, but we found in preliminary experiments that pre-training on this dataset results in lower performance. Nevertheless, it would be exciting to explore the impact of large-scale data cleaning or error correction, and their implications on downstream model performance.

\section*{Ethical Concerns}

\paragraph{Broader Impact.} BERT models remain the most widely used NLP models according to Hugging Face download statistics. First, we investigate the capabilities of modern BERT variants, specifically ModernBERT, in the legal domain. We find that further in domain pre-training yields significant gains, and we expect such results for other domains as well. Second, we hope the resulting artifacts prove to be useful to practitioners and can be used to embed, retrieve or rerank court opinions, or to build closed universe legal tech with specialized in-house models. We see such applications as the main intended use case for our models.  

\paragraph{Intended use case.}\vspace{-0.2em} We recognize that the legal
context is especially sensitive, and caution researchers to think carefully about how they use our released models. For example, efficient legal research could help under-resourced litigants, but it can also facilitate frivolous filings. Similar thoughts hold for other related applications of our models.

\paragraph{Data privacy.}\vspace{-0.2em} All data used in this study is publicly available \citep{harvard-lil_cold_cases, hendersonkrass2022pileoflaw}.

\paragraph{Model Bias.}\vspace{-0.2em} LLMs are known to be biased \cite{abid2021persistent, lucy2021gender}, however we did not investigated such biases in our models. We highlight the exploration of these biases and their mitigation as an important area for future work; and we warrant for caution while using the resulting artifacts in this work and careful validation in downstream applications, especially high-stakes or otherwise sensitive classification or retrieval tasks.


\newpage

\bibliographystyle{ACM-Reference-Format}
\bibliography{sample-base}

@inproceedings{chalkidis-etal-2020-legal,
    title = "{LEGAL}-{BERT}: The Muppets straight out of Law School",
    author = "Chalkidis, Ilias  and
      Fergadiotis, Manos  and
      Malakasiotis, Prodromos  and
      Aletras, Nikolaos  and
      Androutsopoulos, Ion",
    editor = "Cohn, Trevor  and
      He, Yulan  and
      Liu, Yang",
    booktitle = "Findings of the Association for Computational Linguistics: EMNLP 2020",
    month = nov,
    year = "2020",
    address = "Online",
    publisher = "Association for Computational Linguistics",
    url = "https://aclanthology.org/2020.findings-emnlp.261/",
    doi = "10.18653/v1/2020.findings-emnlp.261",
    pages = "2898--2904"
}

@misc{dao2023flashattention2fasterattentionbetter,
      title={FlashAttention-2: Faster Attention with Better Parallelism and Work Partitioning}, 
      author={Tri Dao},
      year={2023},
      eprint={2307.08691},
      archivePrefix={arXiv},
      primaryClass={cs.LG},
      url={https://arxiv.org/abs/2307.08691}, 
}

@misc{modernbert,
      title={Smarter, Better, Faster, Longer: A Modern Bidirectional Encoder for Fast, Memory Efficient, and Long Context Finetuning and Inference}, 
      author={Benjamin Warner and Antoine Chaffin and Benjamin Clavié and Orion Weller and Oskar Hallström and Said Taghadouini and Alexis Gallagher and Raja Biswas and Faisal Ladhak and Tom Aarsen and Nathan Cooper and Griffin Adams and Jeremy Howard and Iacopo Poli},
      year={2024},
      eprint={2412.13663},
      archivePrefix={arXiv},
      primaryClass={cs.CL},
      url={https://arxiv.org/abs/2412.13663}, 
}

@misc{grattafiori2024llama3herdmodels,
      title={The Llama 3 Herd of Models}, 
      author={Aaron Grattafiori and Abhimanyu Dubey and Abhinav Jauhri and Abhinav Pandey and Abhishek Kadian and Ahmad Al-Dahle and Aiesha Letman and Akhil Mathur and Alan Schelten and Alex Vaughan and et al.},
      year={2024},
      eprint={2407.21783},
      archivePrefix={arXiv},
      primaryClass={cs.AI},
      url={https://arxiv.org/abs/2407.21783}, 
}

@misc{webersinke2022climatebertpretrainedlanguagemodel,
      title={ClimateBert: A Pretrained Language Model for Climate-Related Text}, 
      author={Nicolas Webersinke and Mathias Kraus and Julia Anna Bingler and Markus Leippold},
      year={2022},
      eprint={2110.12010},
      archivePrefix={arXiv},
      primaryClass={cs.CL},
      url={https://arxiv.org/abs/2110.12010}, 
}

@inproceedings{beltagy-etal-2019-scibert,
    title = "{S}ci{BERT}: A Pretrained Language Model for Scientific Text",
    author = "Beltagy, Iz  and
      Lo, Kyle  and
      Cohan, Arman",
    editor = "Inui, Kentaro  and
      Jiang, Jing  and
      Ng, Vincent  and
      Wan, Xiaojun",
    booktitle = "Proceedings of the 2019 Conference on Empirical Methods in Natural Language Processing and the 9th International Joint Conference on Natural Language Processing (EMNLP-IJCNLP)",
    month = nov,
    year = "2019",
    address = "Hong Kong, China",
    publisher = "Association for Computational Linguistics",
    url = "https://aclanthology.org/D19-1371/",
    doi = "10.18653/v1/D19-1371",
    pages = "3615--3620"
}

@misc{hendersonkrass2022pileoflaw,
  url = {https://arxiv.org/abs/2207.00220},
  author = {Henderson, Peter and Krass, Mark S. and Zheng, Lucia and Guha, Neel and Manning, Christopher D. and Jurafsky, Dan and Ho, Daniel E.},
  title = {Pile of Law: Learning Responsible Data Filtering from the Law and a 256GB Open-Source Legal Dataset},
  publisher = {arXiv},
  year = {2022}
}

@inproceedings{mahari-etal-2023-law,
    title = "The Law and {NLP}: Bridging Disciplinary Disconnects",
    author = "Mahari, Robert  and
      Stammbach, Dominik  and
      Ash, Elliott  and
      Pentland, Alex",
    editor = "Bouamor, Houda  and
      Pino, Juan  and
      Bali, Kalika",
    booktitle = "Findings of the Association for Computational Linguistics: EMNLP 2023",
    month = dec,
    year = "2023",
    address = "Singapore",
    publisher = "Association for Computational Linguistics",
    url = "https://aclanthology.org/2023.findings-emnlp.224/",
    doi = "10.18653/v1/2023.findings-emnlp.224",
    pages = "3445--3454"
}

@misc{spaeth2020supreme,
  author = {Harold J. Spaeth and Lee Epstein and Jeffrey A. Segal and Andrew D. Martin and Theodore J. Ruger and Sara C. Benesh},
  title = {Supreme Court Database, Version 2020 Release 01},
  year = {2020},
  howpublished = {Washington University Law},
  note = {Accessed: 2025-02-01}
}

@misc{ccap,
    title={Cleaned Caselaw Access Project},
    author={Enrico Shippole, Aran Komatsuzaki},
    howpublished={\url{https://huggingface.co/datasets/TeraflopAI/Caselaw_Access_Project}},
    year={2024}
}

@misc{harvard-lil_cold_cases,
  author       = {Harvard Library Innovation Lab},
  title        = {Cold Cases Dataset},
  year         = {2024},
  url          = {https://huggingface.co/datasets/harvard-lil/cold-cases},
  note         = {Accessed: 2025-02-01}
}

@misc{Feldman2018,
  author = {Adam Feldman},
  title = {Empirical SCOTUS: An opinion is worth at least a thousand words (Corrected)},
  year = {2018},
  month = {April},
  day = {3},
  url = {https://www.scotusblog.com/2018/04/empirical-scotus-an-opinion-is-worth-at-least-a-thousand-words/},
  note = {SCOTUSblog, Accessed: 2025-02-01}
}

@article{SU2024127063,
title = {RoFormer: Enhanced transformer with Rotary Position Embedding},
journal = {Neurocomputing},
volume = {568},
pages = {127063},
year = {2024},
issn = {0925-2312},
doi = {https://doi.org/10.1016/j.neucom.2023.127063},
url = {https://www.sciencedirect.com/science/article/pii/S0925231223011864},
author = {Jianlin Su and Murtadha Ahmed and Yu Lu and Shengfeng Pan and Wen Bo and Yunfeng Liu}
}

@inproceedings{mahari-etal-2024-lepard,
    title = "{L}e{P}a{RD}: A Large-Scale Dataset of Judicial Citations to Precedent",
    author = "Mahari, Robert  and
      Stammbach, Dominik  and
      Ash, Elliott  and
      Pentland, Alex",
    editor = "Ku, Lun-Wei  and
      Martins, Andre  and
      Srikumar, Vivek",
    booktitle = "Proceedings of the 62nd Annual Meeting of the Association for Computational Linguistics (Volume 1: Long Papers)",
    month = aug,
    year = "2024",
    address = "Bangkok, Thailand",
    publisher = "Association for Computational Linguistics",
    url = "https://aclanthology.org/2024.acl-long.532/",
    doi = "10.18653/v1/2024.acl-long.532",
    pages = "9863--9877"
}

@inproceedings{portes2023MosaicBERT,
author = {Portes, Jacob and Trott, Alex and Havens, Sam and King, Daniel and Venigalla, Abhinav and Nadeem, Moin and Sardana, Nikhil and Khudia, Daya and Frankle, Jonathan},
title = {MosaicBERT: a bidirectional encoder optimized for fast pretraining},
year = {2023},
publisher = {Curran Associates Inc.},
address = {Red Hook, NY, USA},
booktitle = {Proceedings of the 37th International Conference on Neural Information Processing Systems},
articleno = {137},
numpages = {25},
location = {New Orleans, LA, USA},
series = {NIPS '23}
}

@misc{niklaus2025lawinstructresourcestudyinglanguage,
      title={LawInstruct: A Resource for Studying Language Model Adaptation to the Legal Domain}, 
      author={Joel Niklaus and Lucia Zheng and Arya D. McCarthy and Christopher Hahn and Brian M. Rosen and Peter Henderson and Daniel E. Ho and Garrett Honke and Percy Liang and Christopher Manning},
      year={2025},
      eprint={2404.02127},
      archivePrefix={arXiv},
      primaryClass={cs.CL},
      url={https://arxiv.org/abs/2404.02127}, 
}

@misc{zheng2021doespretraininghelpassessing,
      title={When Does Pretraining Help? Assessing Self-Supervised Learning for Law and the CaseHOLD Dataset}, 
      author={Lucia Zheng and Neel Guha and Brandon R. Anderson and Peter Henderson and Daniel E. Ho},
      year={2021},
      eprint={2104.08671},
      archivePrefix={arXiv},
      primaryClass={cs.CL},
      url={https://arxiv.org/abs/2104.08671}, 
}

@inproceedings{bert,
    title = "{BERT}: Pre-training of Deep Bidirectional Transformers for Language Understanding",
    author = "Devlin, Jacob  and
      Chang, Ming-Wei  and
      Lee, Kenton  and
      Toutanova, Kristina",
    editor = "Burstein, Jill  and
      Doran, Christy  and
      Solorio, Thamar",
    booktitle = "Proceedings of the 2019 Conference of the North {A}merican Chapter of the Association for Computational Linguistics: Human Language Technologies, Volume 1 (Long and Short Papers)",
    month = jun,
    year = "2019",
    address = "Minneapolis, Minnesota",
    publisher = "Association for Computational Linguistics",
    url = "https://aclanthology.org/N19-1423/",
    doi = "10.18653/v1/N19-1423",
    pages = "4171--4186"
}

@inproceedings{roberta,
    title = "A Robustly Optimized {BERT} Pre-training Approach with Post-training",
    author = "Zhuang, Liu  and
      Wayne, Lin  and
      Ya, Shi  and
      Jun, Zhao",
    editor = "Li, Sheng  and
      Sun, Maosong  and
      Liu, Yang  and
      Wu, Hua  and
      Liu, Kang  and
      Che, Wanxiang  and
      He, Shizhu  and
      Rao, Gaoqi",
    booktitle = "Proceedings of the 20th Chinese National Conference on Computational Linguistics",
    month = aug,
    year = "2021",
    address = "Huhhot, China",
    publisher = "Chinese Information Processing Society of China",
    url = "https://aclanthology.org/2021.ccl-1.108/",
    pages = "1218--1227",
    language = "eng"
}

@inproceedings{lexglue,
    title = "{L}ex{GLUE}: A Benchmark Dataset for Legal Language Understanding in {E}nglish",
    author = "Chalkidis, Ilias  and
      Jana, Abhik  and
      Hartung, Dirk  and
      Bommarito, Michael  and
      Androutsopoulos, Ion  and
      Katz, Daniel  and
      Aletras, Nikolaos",
    editor = "Muresan, Smaranda  and
      Nakov, Preslav  and
      Villavicencio, Aline",
    booktitle = "Proceedings of the 60th Annual Meeting of the Association for Computational Linguistics (Volume 1: Long Papers)",
    month = may,
    year = "2022",
    address = "Dublin, Ireland",
    publisher = "Association for Computational Linguistics",
    url = "https://aclanthology.org/2022.acl-long.297/",
    doi = "10.18653/v1/2022.acl-long.297",
    pages = "4310--4330"
}

@misc{dominguezolmedo2024lawmapowerspecializationlegal,
      title={Lawma: The Power of Specialization for Legal Tasks}, 
      author={Ricardo Dominguez-Olmedo and Vedant Nanda and Rediet Abebe and Stefan Bechtold and Christoph Engel and Jens Frankenreiter and Krishna Gummadi and Moritz Hardt and Michael Livermore},
      year={2024},
      eprint={2407.16615},
      archivePrefix={arXiv},
      primaryClass={cs.CL},
      url={https://arxiv.org/abs/2407.16615}, 
}

@inproceedings{abid2021persistent, 
    title={Persistent anti-muslim bias in large language models},
    author={Abid, Abubakar and Farooqi, Maheen and Zou, James},
    booktitle={Proceedings of the 2021 AAAI/ACM Conference on AI, Ethics, and Society},
    pages={298--306},
    year={2021},
    url={https://dl.acm.org/doi/abs/10.1145/3461702.3462624},
}

@inproceedings{reimers-2019-sentence-bert,
    title = "Sentence-BERT: Sentence Embeddings using Siamese BERT-Networks",
    author = "Reimers, Nils and Gurevych, Iryna",
    booktitle = "Proceedings of the 2019 Conference on Empirical Methods in Natural Language Processing",
    month = "11",
    year = "2019",
    publisher = "Association for Computational Linguistics",
    url = "https://arxiv.org/abs/1908.10084",
}

@misc{tay2022transformermemorydifferentiablesearch,
      title={Transformer Memory as a Differentiable Search Index}, 
      author={Yi Tay and Vinh Q. Tran and Mostafa Dehghani and Jianmo Ni and Dara Bahri and Harsh Mehta and Zhen Qin and Kai Hui and Zhe Zhao and Jai Gupta and Tal Schuster and William W. Cohen and Donald Metzler},
      year={2022},
      eprint={2202.06991},
      archivePrefix={arXiv},
      primaryClass={cs.CL},
      url={https://arxiv.org/abs/2202.06991}, 
}

@inproceedings{chalkidis-etal-2023-lexfiles,
    title = "{L}e{XF}iles and {L}egal{LAMA}: Facilitating {E}nglish Multinational Legal Language Model Development",
    author = "Chalkidis, Ilias  and
      Garneau, Nicolas  and
      Goanta, Catalina  and
      Katz, Daniel  and
      S{\o}gaard, Anders",
    booktitle = "Proceedings of the 61st Annual Meeting of the Association for Computational Linguistics (Volume 1: Long Papers)",
    month = jul,
    year = "2023",
    address = "Toronto, Canada",
    publisher = "Association for Computational Linguistics",
    url = "https://aclanthology.org/2023.acl-long.865",
    pages = "15513--15535"
}

@inproceedings{Zheng_2025,
author = {Zheng, Lucia and Guha, Neel and Arifov, Javokhir and Zhang, Sarah and Skreta, Michal and Manning, Christopher D. and Henderson, Peter and Ho, Daniel E.},
title = {A Reasoning-Focused Legal Retrieval Benchmark},
year = {2025},
isbn = {9798400714214},
publisher = {Association for Computing Machinery},
address = {New York, NY, USA},
url = {https://doi.org/10.1145/3709025.3712219},
doi = {10.1145/3709025.3712219},
booktitle = {Proceedings of the 2025 Symposium on Computer Science and Law},
pages = {169–193},
numpages = {25},
location = {Munich, Germany},
series = {CSLAW '25}
}

@inproceedings{zhong-etal-2020-nlp,
    title = "How Does {NLP} Benefit Legal System: A Summary of Legal Artificial Intelligence",
    author = "Zhong, Haoxi  and
      Xiao, Chaojun  and
      Tu, Cunchao  and
      Zhang, Tianyang  and
      Liu, Zhiyuan  and
      Sun, Maosong",
    editor = "Jurafsky, Dan  and
      Chai, Joyce  and
      Schluter, Natalie  and
      Tetreault, Joel",
    booktitle = "Proceedings of the 58th Annual Meeting of the Association for Computational Linguistics",
    month = jul,
    year = "2020",
    address = "Online",
    publisher = "Association for Computational Linguistics",
    url = "https://aclanthology.org/2020.acl-main.466/",
    doi = "10.18653/v1/2020.acl-main.466",
    pages = "5218--5230"
}

@inproceedings{lucy2021gender, 
    title={Gender and representation bias in {GPT}-3 generated stories},
    author={Lucy, Li and Bamman, David},
    booktitle={Proceedings of the Third Workshop on Narrative Understanding},
    pages={48--55},
    year={2021},
    url={https://aclanthology.org/2021.nuse-1.5/},
}

@misc{niklaus2024multilegalpile689gbmultilinguallegal,
      title={MultiLegalPile: A 689GB Multilingual Legal Corpus}, 
      author={Joel Niklaus and Veton Matoshi and Matthias Stürmer and Ilias Chalkidis and Daniel E. Ho},
      year={2024},
      eprint={2306.02069},
      archivePrefix={arXiv},
      primaryClass={cs.CL},
      url={https://arxiv.org/abs/2306.02069}, 
}

@misc{stammbach2026legalretrievalpublicdefenders,
      title={Legal Retrieval for Public Defenders}, 
      author={Dominik Stammbach and Kylie Zhang and Patty Liu and Nimra Nadeem and Inyoung Cheong and Lucia Zheng and Peter Henderson},
      year={2026},
      eprint={2601.14348},
      archivePrefix={arXiv},
      primaryClass={cs.IR},
      url={https://arxiv.org/abs/2601.14348}, 
}

@misc{cheong2026aiaugmentaccessjustice,
      title={How Can AI Augment Access to Justice? Public Defenders' Perspectives on AI Adoption}, 
      author={Inyoung Cheong and Patty Liu and Dominik Stammbach and Peter Henderson},
      year={2026},
      eprint={2510.22933},
      archivePrefix={arXiv},
      primaryClass={cs.CY},
      url={https://arxiv.org/abs/2510.22933}, 
}

@article{privilege,
    author = {Adam Paine and Robert M. Travisano},
    journal = {Epstein Becker Green}, 
    title = {Discovery Pitfalls in the Age of AI},
    howpublished = {\url{https://techcrunch.com/2025/07/25/sam-altman-warns-theres-no-legal-confidentiality-when-using-chatgpt-as-a-therapist/}},
    year = {2025}, 
    month = {September}}

@inproceedings{cheong2024not,
  title={(A) I am not a lawyer, but...: engaging legal experts towards responsible LLM policies for legal advice},
  author={Cheong, Inyoung and Xia, King and Feng, KJ Kevin and Chen, Quan Ze and Zhang, Amy X},
  booktitle={Proceedings of the 2024 ACM Conference on Fairness, Accountability, and Transparency},
  pages={2454--2469},
  year={2024}
}

@inproceedings{translating_legalese,
author = {Ash, Elliott and Kesari, Aniket and Naidu, Suresh and Song, Lena and Stammbach, Dominik},
title = {Translating Legalese: Enhancing Public Understanding of Court Opinions with Legal Summarizers},
year = {2024},
isbn = {9798400703331},
publisher = {Association for Computing Machinery},
address = {New York, NY, USA},
url = {https://doi.org/10.1145/3614407.3643700},
doi = {10.1145/3614407.3643700},
booktitle = {Proceedings of the Symposium on Computer Science and Law},
pages = {136–157},
numpages = {22},
location = {Boston, MA, USA},
series = {CSLAW '24}
}

\appendix
\onecolumn

\section{Appendix}\label{appendix:model_training}

\begin{figure}[h!]
    \centering
    \caption{Loss curves during pre-training: number of steps on the x-axis, loss on the y-axis. Top row displays curves initialized from existing ModernBERT checkpoint. Bottom row displays curves if trained from scratch. First column shows base models, second column shows large models (all curves averaged using a moving window average of 20 batches).}
    \label{fig:loss-curves}
    
    \includegraphics[width=0.38\linewidth]{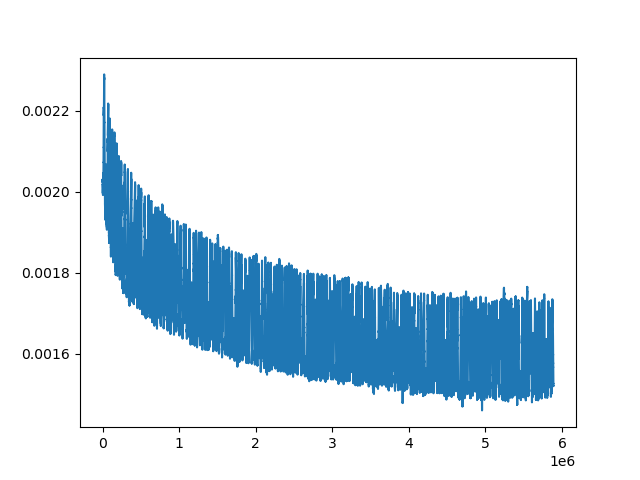}
    \includegraphics[width=0.38\linewidth]{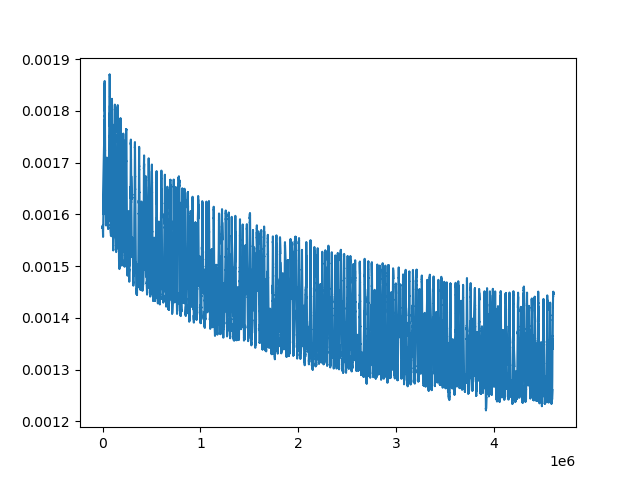}

    \includegraphics[width=0.38\linewidth]{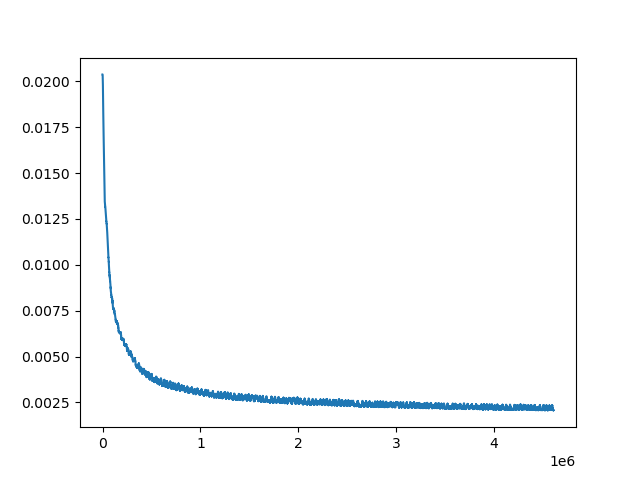}
    \includegraphics[width=0.38\linewidth]{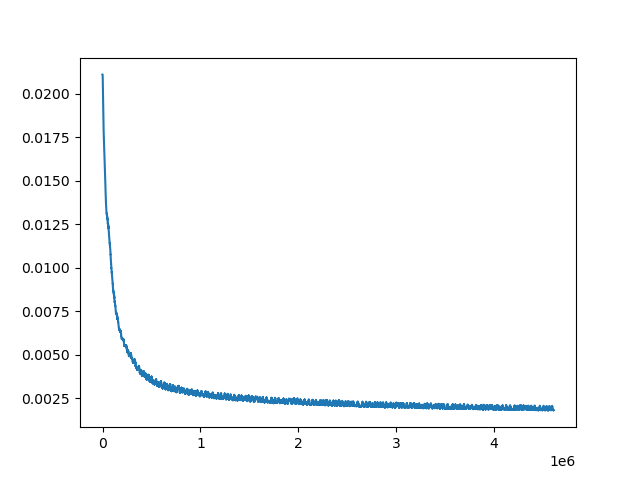}

\end{figure}

All our model checkpoints and our replication package are available under the same license as ModernBERT (Apache License 2.0) \citep{modernbert} and are available on Hugging Face: \href{https://huggingface.co/ai-law-society-lab/CaseLawModernBERT-base}{https://huggingface.co/ai-law-society-lab/CaseLawModernBERT-base} and 
\href{https://huggingface.co/ai-law-society-lab/CaseLawModernBERT-large}{https://huggingface.co/ai-law-society-lab/CaseLawModernBERT-large}. Pre-training runs took 60 to 72 hours on a single GPU (base and large runs). All fine-tuning runs were 2x faster than the times reported in Table 7 in \citep{lexglue}. This is consistent with the speed gains reported in Section 4.2 in \citep{modernbert}. We show pre-training loss curves in Figure \ref{fig:loss-curves}.

\begin{table*}
    \centering
    \caption{Probing results of all examined models following the evaluation setup outlined in \citep{chalkidis-etal-2023-lexfiles}. The released models in this work are further pre-trained on only US court opinions. We note that the resulting models achieve the highest P@1 and MRR for US terms and US crimes -- and for the average score. At the same time, we observe that P@1 and MRR for e.g., European Court of Human Rights terms decrease after further pre-training on US court opinions.
    }
    \label{tab:probing_results}
    
    \small
    
    \begin{tabular}{l c c c c c c c c c c c c}
    Task & \multicolumn{2}{c}{ModernBERT-base} & \multicolumn{2}{c}{From scratch base }  & \multicolumn{2}{c}{LegalModernBERT-base} & \multicolumn{2}{c}{ModernBERT-Large} & \multicolumn{2}{c}{From scratch large} & \multicolumn{2}{c}{LegalModernBERT-large}    
    \\
    & P@1 & MRR & P@1 & MRR & P@1 & MRR & P@1 & MRR & P@1 & MRR & P@1 & MRR
    \\ \hline
cjeu terms & 0.29 & 0.41 & 0.23 & 0.36 & 0.27 & 0.4 & 0.38 & 0.5 & 0.23 & 0.36 & 0.33 & 0.45 \\
contract types & 0.18 & 0.34 & 0.34 & 0.52 & 0.21 & 0.38 & 0.27 & 0.47 & 0.31 & 0.46 & 0.41 & 0.56  \\
contract sections & 0.07 & 0.23 & 0.59 & 0.71 & 0.26 & 0.44 & 0.16 & 0.32 & 0.51 & 0.66 & 0.4 & 0.55  \\
canadian crimes & 0.16 & 0.2 & 0.18 & 0.22 & 0.19 & 0.23 & 0.22 & 0.25 & 0.18 & 0.23 & 0.21 & 0.25 \\
us terms & 0.38 & 0.48 & 0.56 & 0.65 & 0.5 & 0.6 & 0.53 & 0.62 & 0.55 & 0.64 & \textbf{0.62} & \textbf{0.7} \\
us crimes & 0.39 & 0.5 & 0.48 & 0.59 & 0.45 & 0.57 & 0.52 & 0.63 & 0.49 & 0.6 & \textbf{0.55} & \textbf{0.66} \\
ecthr articles & 0.24 & 0.38 & 0.26 & 0.39 & 0.26 & 0.39 & 0.29 & 0.45 & 0.24 & 0.38 & 0.27 & 0.4 \\
ecthr terms & 0.39 & 0.45 & 0.34 & 0.41 & 0.38 & 0.44 & 0.52 & 0.58 & 0.36 & 0.42 & 0.47 & 0.54 \\
Average & 0.26 & 0.37 & 0.37 & 0.48 & 0.31 & 0.43 & 0.36 & 0.48 & 0.36 & 0.47 & \textbf{0.41} & \textbf{0.52}\\ \hline
    \end{tabular}
\end{table*}

\begin{table*}[t!]
    \centering
    \caption{Development set results on two LexGLUE tasks. Results are average across 5 runs with 95\% confidence intervals. ** denotes statistically significant results.
    } 
    \label{tab:dev_main_results}

    \small
    \begin{tabular}{l l c c }
        Dataset & Model & Micro-F1 & Macro-F1 \\ \hline
        \multirow{4}{*}{LexGLUE/SCOTUS base} & Best Reported in \cite{lexglue} & 74.0 & 81.3 \\ 
        & ModernBERT-base & 73.6 $\pm$ 1.7 & 80.0 $\pm$ 0.5 \\
        & LegalModernBERT-base from scratch & 72.4 $\pm$ 0.2 & 80.0 $\pm$ 0.4 \\
        & LegalModernBERT-base & 75.5 $\pm$ 1.8 & 81.2 $\pm$ 0.9 \\ \hdashline
        \multirow{4}{*}{LexGLUE/SCOTUS large} &  RoBERTa-large \cite{lexglue} & 56.9 & 74.6 \\
        & ModernBERT-large & 75.4 $\pm$ 0.6 & 81.5 $\pm$ 0.7 \\
        & LegalModernBERT-large from scratch & 73.4 $\pm$ 0.4 & 79.8 $\pm$ 0.4 \\
        & LegalModernBERT-large & \textbf{76.2} $\pm$ 1.8 & \textbf{81.8} $\pm$ 0.6 \\ \hline
        \multirow{4}{*}{LexGLUE/CaseHold base} & Best Reported in \cite{lexglue} & 77.4 & 77.4 \\ 
        & ModernBERT & 76.4 $\pm$ 0.1 & 76.4 $\pm$ 0.1 \\
        & LegalModernBERT from scratch & 77.2 $\pm$ 0.3 & 77.2 $\pm$ 0.3 \\
        & LegalModernBERT & 77.7 $\pm$ 0.1 & 77.7 $\pm$ 0.1 \\ \hdashline
        \multirow{4}{*}{LexGLUE/CaseHold large} & RoBERTa-large \cite{lexglue} & 76.8 & 76.8 \\ 
        & ModernBERT-large & 78.6 $\pm$ 0.4 & 78.6 $\pm$ 0.4 \\
        & LegalModernBERT-large from scratch & 77.0 $\pm$ 0.3 & 77.0 $\pm$ 0.3 \\
        & LegalModernBERT-large & \textbf{79.6 $\pm$ 0.1**} & \textbf{79.6 $\pm$ 0.1**} \\ \hline
        
    \end{tabular}
\end{table*}

\begin{table}[]
    \centering
    \caption{Results on LePaRD using zero-shot bi-encoders without any further fine-tuning (embedding queries and targets with mean pooling and retrieving the targets with the highest cosine similarity to the query).}
    \label{tab:zero-shot-lepard}

    \small
    \begin{tabular}{l c c}
    
    Model & rc@1 & rc@10 \\ \hline
    ModernBERT-base & 2.8 & 6.9 \\
    LegalModernBERT-base from scratch & \textbf{4.1} & \textbf{11.1} \\
    LegalModernBERT-base & 3.3 & 8.9 \\ \hdashline
    ModernBERT-large & 3.1 & 8.1 \\
    LegalModernBERT-large from scratch & \textbf{4.2} & \textbf{11.3} \\
    LegalModernBERT-large & 3.6 & 10.0 \\ \hline
    \end{tabular}
\end{table}

\begin{table}[]
    \centering
    \caption{Results on LePard (classification setting using the top 10K passages). Reported results are the average of 5 runs with 95\% confidence intervals. ** denotes statistically significant results.}
    \label{tab:lepard_classification}
    
    \small
    \begin{tabular}{l  c c}
    
    Model &  \multicolumn{2}{c}{Test}   \\
    &  rc@1 & rc@10 \\ \hline
    Best reported in \cite{mahari-etal-2024-lepard} & 38.0 & 81.2 \\ \hdashline
    ModernBERT-base & 38.03 $\pm$ 0.0 & 83.16 $\pm$ 0.2 \\
    LegalModernBERT-base from scratch & 38.44 ± 0.1 & 85.23 ± 0.1 \\

    LegalModernBERT-base & 38.49 ± 0.1 & 84.00 ± 0.1 \\ \hdashline
    ModernBERT-large & 39.25 ± 0.1 & 86.52 ± 0.1 \\
    LegalModernBERT-large from scratch & 38.01 ± 0.1 & 86.1 ± 0.0 \\
    LegalModernBERT-large & \textbf{39.54 ± 0.1**} & \textbf{87.37 ± 0.2} \\ \hline
    \end{tabular}
\end{table}

\end{document}